\documentclass[wcp]{jmlr}

\usepackage{longtable}
\usepackage{booktabs}

\usepackage{algorithmic}
\usepackage{algorithm} 
\let\svthefootnote\thefootnote
\newcommand\blankfootnote[1]{%
  \let\thefootnote\relax\footnotetext{#1}%
  \let\thefootnote\svthefootnote%
}

\usepackage[utf8x]{inputenc}

\title{Latent Multi-view Semi-Supervised Classification}

  \author{\Name{Xiaofan Bo}\\
  \addr  Glasgow college, University of Electronic Science and Technology of China
  \AND
  \Name{Zhao Kang} \Email{Zkang@uestc.edu.cn}\\
  \addr  School of Computer Science and Engineering, University of Electronic Science and Technology of China
    \AND
  \Name{Zhitong Zhao} \\
  \addr  School of Computer Science and Engineering, University of Electronic Science and Technology of China
    \AND
  \Name{Yuanzhang Su}\Email{syz@uestc.edu.cn}\\
  \addr  School of Foreign Languages, University of Electronic Science and Technology of China
    \AND
  \Name{Wenyu Chen}\Email{cwy@uestc.edu.cn}\\
  \addr  School of Computer Science and Engineering, University of Electronic Science and Technology of China
 }

\begin{document}

\maketitle

\begin{abstract}
To explore underlying complementary information from multiple views, in this paper, we propose a novel Latent Multi-view Semi-Supervised Classification (LMSSC) method. Unlike most existing multi-view semi-supervised classification methods that learn the graph using original features, our method seeks an underlying latent representation and performs graph learning and label propagation based on the learned latent representation. With the complementarity of multiple views, the latent representation could depict the data more comprehensively than every single view individually, accordingly making the graph more accurate and robust as well. Finally, LMSSC integrates latent representation learning, graph construction, and label propagation into a unified framework, which makes each subtask optimized. Experimental results on real-world benchmark datasets validate the effectiveness of our proposed method.
\end{abstract}
\begin{keywords}
Semi-supervised classification; Multi-view learning; Latent space
\end{keywords}

\section{Introduction}
\label{sec:intro}
Thanks to its ability to take advantage of abundant unlabeled data, semi-supervised classification has been widely used for numerous problems \cite{chapelle2009semi}. A great number of semi-supervised classification methods have been developed in the past decades \cite{zhu2009introduction}. Among them, the graph-based semi-supervised classification technique has achieved state-of-the-art performance. Given a dataset with a limited number of initial labels, it labels the unlabeled ones according to the propagation of pairwise similarity. In particular, it consists of two steps. First, a graph is constructed based on certain similarity metrics. Each data point is represented by a node on the graph and the weight denotes the similarity between two points. Second, the labels of unlabeled points are inferred by label propagation. Therefore, numerous methods focus on either building graphs \cite{jebara2009graph,cheng2009sparsity,zhuang2012non,li2015learning,li2017multi}, or designing effective label propagation schemes \cite{zhou2004learning,wang2008label,zhu2003semi}. 

Although existing techniques have achieved promising performance in various real-world applications, they are limited in three aspects. First, they mainly deal with single-view data. Nowadays, data is often represented by multiple views brought by various sensors and feature descriptors \cite{tao2017ensemble,kang2019multiple}. Moreover, leveraging multi-view data can boost the performance of single-view methods, due to the complementarity of heterogeneous information. Many machine learning, data mining, pattern recognition, and computer vision tasks have benefited from multi-view data \cite{xu2013survey,zhao2017multi,fu2015transductive}. As a result, it is paramount to develop multi-view semi-supervised classification techniques. Though some methods have been developed to tackle multi-view data, they suffer other drawbacks.

Second, most graphs are constructed from the original data. Real-world data is often contaminated by various noise or outliers. Thus, the resulted graph may not be accurate to reflect the underlying relationships between data samples \cite{kang2019low,kang2019Clustering}. As a matter of fact, many researchers have shown that graph quality is crucial to the performance of subsequent tasks \cite{nie2017multi,kang2017twin,kang2019similarity,kang2018unified}. Therefore, how to construct a robust and reliable graph is vital. For multi-view data, this is more challenging due to its heterogeneity nature.

Third, existing methods usually take graph construction and label propagation as two separate steps. Consequently, they are not jointly optimized. In particular, the resulted graph might not be optimal for a subsequent task. Furthermore, it is desired to exploit the partial label information to guide the graph construction process. Existing methods often fail to make use of this information.

Confronted with the aforementioned limitations, we propose a novel multi-view learning method, named as Latent Multi-view Semi-Supervised Classification (LMSSC). It is composed of three components. The first component extracts view-specific and shared latent factors from all the views. In specific, the shared latent factor can be treated as a view-independent data representation. Based on it, we can construct a common graph for all views. This is reasonable since points in different views indeed represent the same set of objects.
Unlike traditional fixed graph, we learn it from data and update it iteratively according to the result of classification.
After the graph is obtained, the label propagation function is updated accordingly. In this way, the latent factor, the graph, and the label propagation are jointly optimized.

The main merits of this paper can be summarized as follows:
\begin{itemize}
  \item {We propose a novel multi-view semi-supervised classification method, LMSSC. It integrates common representation learning, graph construction, and label prediction into a unified framework.}
  \item {The view consistency is ensured by extracting shared latent factor from multi-view data. A robust graph is built on the latent factor with structure guarantee. }
  \item{We conduct extensive experiments on benchmark datasets. Compared with the representative single-view and multi-view semi-supervised classification methods, our proposed method demonstrates its superiority.}
\end{itemize}
\textbf{Notations} For a matrix $M$, we express its $i$-th row and $j$-th column element as $M_{ij}$. The Frobenius norm of matrix $M$ is defined as the square root of summation of the square of every single element, i.e., $\left \| M \right \|_{F}=\sqrt{\sum _{ij}M_{ij}^{2}}$, while the $\ell_{2}$-norm of vector $m$ is denoted by $\left \| m\right \|_{2}=\sqrt{m^{T} \cdot m}$. The trace operator is expressed as $Tr(\cdot)$. $\vec{1}$ represents a column vector whose elements are all ones.

\section{Related Work}

Recently, graph-based semi-supervised classification has attracted a lot of attention. For example, Zhu et al. \cite{zhu2003semi} designed a semi-supervised classification algorithm based on Gaussian Field and Harmonic function (GFHF). It takes advantage of the harmonic property of Gaussian random field over the graph. Though it has gained huge popularity, its performance heavily depends on the input graph. Later, Nie et al. \cite{nie2011unsupervised} design a semi-supervised classification method by minimizing the $\ell_1$-norm of spectral embedding; Sparse \cite{he2011nonnegative,yan2009semi} and low-rank graphs \cite{zhuang2012non,kang2018self,kang2019robust} have also been proposed. These methods all focus on single-view data and cannot make full use of the multi-view information.

Some semi-supervised classification techniques have been extended to the multi-view setting. Nie et al. propose Auto-weighted Multiple Graph Learning (AMGL) \cite{nie2016parameter}. For multi-view data set $X=\left \{ X^{v} \right \}, v=1,2,3,\cdots, V $, where $V$ represents the number of views. Each view contains $N$ samples denoted as $\left \{ x_{1}^{v}, x_{2}^{v},..., x_{N}^{v}\right \}\in \mathcal{R}^{d_{(v)}\times N}$, where $d_{(v)}$ denotes the feature number of the $v$-th view. Given graph matrix $S=\{s_{ij}\}\in\mathcal{R}^{N\times N}$, the corresponding degree matrix $D$ $(d_{ii}=\sum_{j=1}^N s_{ij})$ is available. Then, $L=D-S$ is the so-called Laplacian matrix. Without loss of generality, all the points are rearranged and the front $l(l<N)$ points are labeled. AMGL solves
\begin{equation}
\min_F \sum\limits_{v=1}^V  w^{v}Tr(F^TL^{v}F) \quad s.t.\quad f_i=y_i, \quad \forall i=1, 2,\cdots, l,
\end{equation}
where the $v$-th view weight $w^{v}=1\Big/ \left(2\sqrt{Tr(F^TL^{(v)}F)}\right)$ is updated iteratively, $F=[f_1,\cdots,f_n]^T\in\mathcal{R}^{N\times c}$ is the class indicator matrix for $c$ classes and $y_i$ is the given indicator vector for the $i$-th point. $y_{ij}=1$ only if the $i$-th point belongs to the $j$-th class, otherwise $y_{ij}=0$. This approach suffers the graph construction issue.

To address the above problem, Multi-view Learning with Adaptive Neighbours (MLAN) \cite{nie2017multi} is further developed. It learns the graph based on adaptive neighbours. Basically, it solves the following problem
\begin{equation}
\min_S \sum\limits_{v=1}^V w^{v}\sum_{ij}\|x_i^{v}-x_j^{v}\|_2^2s_{ij}+\alpha\|S\|_F^2 \quad s.t.\quad s_i^T\vec{1}=1, \quad 0\leq s_{ij}\leq1,
\end{equation}
where $\alpha>0$ is a trade-off parameter and view weight is updated according to $w^{v}=1\Big/ \left(2\sqrt{\sum_{ij}\|x_i^{v}-x_j^{v}\|_2^2 s_{ij}}\right)$. The distance between $x_{i}$ and $x_{j}$ and the similarity are negatively correlated since adjacent samples have larger similarity. In traditional graph construction approaches, the neighbours are pre-determined and the similarity is fixed. Differing from this, MLAN assigns adaptive neighbours to each data point. In other words, the $k$ nearest neighbours of any $x_i$ are not steady and they change in every iteration. As a result, this strategy always shows better performance than previous heuristic approaches. Despite its promising performance, the graph is built on the raw data which can be easily contaminated by noise or outliers. In addition, the weight assignment is also quite arbitrary, which could lead to an unsatisfied solution.
 
\section{Proposed Methodology}
As demonstrated above, there exist two key problems that should be solved for multi-view semi-supervised classification. First, how to construct a robust graph based on multi-view data? Second, how to combine the graph construction with label propagation process? To address these problems, we propose a latent multi-view semi-supervised classification (LMSSC) method. 
\subsection{Formulation}

 The integrated formulation is as follows:
\begin{equation}
\underset{W,H,S,F}{\min}\Phi (X,W,H)+\beta \Omega (H,L,S)+\gamma \Theta(L,Y, F),
\label{unified}
\end{equation} 
where $W=\left \{ W^{v}\in \mathcal{R}^{d_{(v)}\times r}, v=1,2,...,V \right \}$ and $H\in \mathcal{R}^{r\times N}$ are latent factors extracted from multi-view data $X$ and $r$ is the dimension of latent representation. $S$ is the similarity graph shared by all views. $L$ is the Laplacian matrix of $S$ and $Y$ is a prior label indicator matrix. $F$ is the label indicator matrix that we aim to predict. $\beta$ and $\gamma$ are trade-off parameters. Next, we will discuss each term in problem (\ref{unified}) in detail.

\subsection{Factors in Latent Space}
In order to build a robust graph, we learn a shared latent factor from all views. Since points in different views indeed represent the same set of objects, the shared latent factor is considered to be view-independent features. 

Specifically, we decompose the samples in $v$-th view as $X^{v}=W^{v}H$, where $W^v$ can be treated as view-specific factor and $H$ is the latent representation shared by all views. Consequently, the first term in problem (\ref{unified}) can be formulated as
\begin{equation}
\Phi (X,W,H)=\sum_{v=1}^{V}\left \| X^{v}-W^{v}H \right \|_{F}^{2}\quad s.t.\quad W^{v}\geq 0.
\label{1st}
\end{equation}
Each $h_i$ denotes a new representation of object $i$.

\subsection{Graph Construction}
With the shared latent representation $H$, we can build a graph on it. Hence, this graph is also shared by all views. Since adaptive neighbours strategy can capture the local manifold structure of data \cite{nie2017multi}, we utilize this approach to construct similarity graph $S$. Then, the second term in problem (\ref{unified}) is formulated as follows:

\begin{equation}
\Omega (H,L,S)=\frac{1}{2}\sum_{i,j=1}^{N}\left \| h_{i}-h_{j} \right \|_{2}^{2}s_{ij}+\alpha \left \| S \right \|_{F}^{2} \quad s.t.\quad s_{i}^{T}\mathbf{1}=1,\quad 0\leq s_{ij}\leq 1.
\label{cons}
\end{equation}
Moreover, we have the following equality
\begin{equation}
\frac{1}{2}\sum_{i,j=1}^{N}\left \| h_{i}-h_{j} \right \|_{2}^{2}s_{ij}=Tr(HLH^{T}).\end{equation}
So formulation (\ref{cons}) can be simplified as 
\begin{equation}
\Omega (H,L,S)=Tr(HLH^{T})+\alpha \left \| S \right \|_{F}^{2}\quad s.t.\quad s_{i}^{T}\mathbf{1}=1,\quad 0\leq s_{ij}\leq 1.
\label{2nd}
\end{equation}
Though this graph is built on latent space, there is no guarantee that it would be optimal for subsequent classification. Ideally, graph $S$ should have exact $c$ connected components, i.e., the data points are already identified as $c$ classes. However, the current solution can hardly satisfy such a condition. To tackle this problem, we can resort to the following theorem \cite{mohar1991laplacian}:
\begin{theorem}
The number of connected components $c$ of the graph $S$ is equal to the multiplicity of zero eigenvalue of its Laplacian matrix $L$. 
\end{theorem}
We denote $\sigma _{i}(L)$ as the $i$-th smallest eigenvalue of $L$. Since $L$ is a positive semidefinite matrix, its eigenvalues $\sigma_i(L)\geq0$. Theorem 1 indicates that if $\sum_{i=1}^c \sigma_i=0$, then our requirement can be met. Hence, we can achieve a desired graph by minimizing $\sum_{i=1}^c \sigma_i$. Taking this into account, we can formulate Eq. (\ref{2nd}) as
\begin{equation}
\Omega (H,L,S)=Tr(HLH^{T})+\alpha \left \| S \right \|_{F}^{2}+\gamma\sum_{i=1}^c \sigma_i \quad s.t.\quad s_{i}^{T}\mathbf{1}=1,\quad 0\leq s_{ij}\leq 1.
\label{22nd}
\end{equation}

\subsection{Label Propagation}
In fact, Eq. (\ref{22nd}) is hard to handle due to the involvement of graph structure term. Fortunately, Ky Fan's theorem \cite{fan1949theorem} gives the following equality:
\begin{equation}
\sum\limits_{i=1}^c \sigma_i=\min_{F,F^TF=I} Tr(F^TLF),
\label{fan}
\end{equation}
For semi-supervised learning, $F$ can be decomposed as $F=\left [ F_{l};F_{u} \right ]=\left [ Y_{l};F_{u} \right ]$, where $Y_{l}=\left [ y_{1},y_{2},...,y_{l} \right ]^{T}$ represents the known label matrix and $l(u)$ is the number of labeled(unlabeled) data points. With these notations, the right part of above equation becomes the objective function of semi-supervised classification \cite{nie2017multi,nie2016parameter}. Therefore, the requirement for the graph structure and label propagation are the same in essence. Then, the third term in problem (\ref{unified})  can be expressed as:
\begin{equation}
\Theta(L,Y, F)=Tr(F^TLF)\quad s.t. \quad F_l=Y_l.
\label{3rd}
\end{equation}
\subsection{Unified Objective Function}
Based on Eqs. (\ref{1st}), (\ref{2nd}), and (\ref{3rd}), our objective function (\ref{unified}) can be explicitly written as

\begin{equation}
\begin{split}
\min_{W,H,S,F}& \sum_{v=1}^{V}\left \| X^{v}-W^{v}H \right \|_{F}^{2}+\beta (Tr(HLH^{T})+\alpha \left \| S \right \|_{F}^{2})+\gamma Tr(F^{T}LF)\\
s.t.&\quad W^{v}\geq 0,s_{i}^{T}\mathbf{1}=1,0\leq s_{ij}\leq 1,F_{l}=Y_{l}.
\end{split}
\label{finalobj}
\end{equation}

We can observe that Eq. (\ref{finalobj}) integrates latent representation learning, graph construction, and label prediction into a unified framework. Joint optimization of $H$, $S$, and $F$ will facilitate an overall optimal solution. Furthermore, graph $S$ is built in latent space, thus it is robust to noise and outliers in general.

\section{Optimization}
Eq. (\ref{finalobj}) is not convex with respect to all variables. Thus we solve problem (\ref{finalobj}) based on an alternating strategy, i.e., we solve one variable while considering other variables stationary. \\
$\textbf{A.} \textit{ Update View-Specific Latent Factor }W^{v}$

When $H$, $S$, and $F$ are fixed, problem (\ref{finalobj}) turns into 
\begin{equation}
\min_{W^v}\sum_{v=1}^{V}\left \| X^{v}-W^{v}H \right \|_{F}^{2}\quad s.t.\quad W^{v}\geq 0.
\end{equation}	
This can be solved column-wisely, i.e.,
\begin{equation}
\min_{W_{i,:}^{v}} \|X_{i,:}^{v}-W_{i,:}^{v}H\|_2^2\quad s.t.\quad W_{i,:}^{v}\geq 0.
\label{updatew}
\end{equation}
It is a quadratic programing problem which can be easily solved by many existing packages.

$\textbf{B.} \textit{ Update Shared Latent Factor }H$

We set variables other than $H$ fixed, then we have the following subproblem 
 \begin{equation}
 \min_H \sum_{v=1}^{V}\left \| X^{v}-W^{v}H \right \|_{F}^{2}+\beta Tr(HLH^{T}).
 \end{equation}

By setting the derivative w.r.t. $H$ to zero, we obtain
\begin{equation}
\sum_{v=1}^{V}(W^{v})^{T}W^{v}H+\beta HL=\sum_{v=1}^{V}(W^{v})^{T}X^{v}.
\label{updateh}
\end{equation}
The above equation is the so-called Sylvester equation and can be easily solved by the Bartels-Stewart algorithm \cite{bartels1972solution}. It has a unique solution since $\sum_{v=1}^{V}(W^{v})^{T}W^{v}$ and $-\beta L$ have no common eigenvalues \cite{zhang2017latent}.

$\textbf{C.}\textit{ Update Similarity Graph }S$

After ignoring non-relevant variables, we get 
\begin{equation}\
\min_S \beta (Tr(HLH^{T})+\alpha \left \| S \right \|_{F}^{2})+\gamma Tr(F^{T}LF)\quad s.t.\quad s_{i}^{T}\mathbf{1}=1,0\leq s_{ij}\leq 1.
\label{ss}
\end{equation}
Remember that $L$ is a function of $S$ and $Tr(HLH^{T})=\frac{1}{2}\sum\limits_{i,j=1}^{N}\left \| h_{i}-h_{j} \right \|_{2}^{2}s_{ij}$. Let $d_{ij}^{h}=\left \| h_{i}-h_{j} \right \|_{2}^{2}$
and $d_{ij}^{f}=\left \| f_{i}-f_{j} \right \|_{2}^{2}$, then we can reformulate problem (\ref{ss}) column-wisely as
\begin{equation}
\min_{s_i}\sum_{j=1}^{N}\beta (\frac{1}{2}d_{ij}^{h}s_{ij}+\alpha s_{ij}^{2})+\frac{1}{2}\gamma d_{ij}^{f}s_{ij}\quad s.t.\quad s_{i}^{T}\mathbf{1}=1,0\leq s_{ij}\leq1.
\end{equation} 
Denote $d_{i}\in R^{N\times 1}$ a column vector with $d_{ij}=\beta d_{ij}^{h}+\gamma d_{ij}^{f}$. Then above problem can be simplified as
 \begin{equation}
\min_{s_i} \left \| s_{i}+\frac{1}{4\alpha \beta }d_{i} \right \|_{2}^{2}\quad s.t.\quad  s_{i}^{T}\mathbf{1}=1,\quad0\leq s_{ij}\leq 1.
\label{updates}
\end{equation}
We will show its closed-form solution in the next section.

$\textbf{D.}\textit{ Update Label Indicator }F$

For $F$, the remaining terms are
\begin{equation}
\min_F Tr(F^TLF) \quad s.t.\quad F_{l}=Y_{l}.
\label{upf}
\end{equation}
We first split $L$ into blocks as $L=\begin{bmatrix}
L_{ll} & L_{lu}\\ 
L_{ul} & L_{uu}
\end{bmatrix}.\
$We take the derivative of Eq. (\ref{upf}) in terms of $F$ and set its first-order derivative to zero, that is, 
\begin{equation}\begin{bmatrix}
L_{ll} & L_{lu}\\ 
L_{ul} & L_{uu}
\end{bmatrix}\begin{bmatrix}
Y_{l}\\ F_{u}
\end{bmatrix}=0.\end{equation}
We can see that 
\begin{equation}
F_{u}=-L_{uu}^{-1}L_{ul}Y_{l}.
\label{updatef}
\end{equation}

We iteratively update above variables until maximum 100 times are reached or the relative change of $F$ is less than $10^{-5}$. The complete procedures are shown in Algorithm 1.

Finally, the labels for unlabeled points can be assigned according to the following decision function:
\begin{equation}
y_{i}=\arg \underset{j}{\max  }\hspace{.1cm}  F_{ij},\quad\forall i=l+1,l+2,\cdots, N. \quad\forall j=1,2,\cdots, c.
\label{updatey}
\end{equation}

\begin{algorithm}

\caption{The algorithm of LMSSC}
\label{alg2}
 {\bfseries Input:} Data matrices: $X^{(1)},\cdots, X^{(V)}$, label matrix $Y_l$, parameters $\beta>0$, $\gamma>0$.\\
{\bfseries Initialize:} Random matrix $H$ and $S$, $F_u=0$.\\
 {\bfseries REPEAT}
\begin{algorithmic}[1]
\STATE Calculate $W$ by (\ref{updatew}).
 \STATE Update $H$ by solving (\ref{updateh}).
\STATE Calculate $S$ by solving (\ref{updates}) 
\STATE Update $F$ using (\ref{updatef}).
\end{algorithmic}
\textbf{ UNTIL} {stopping criterion is met.}
\end{algorithm}

\subsection{Determine Value of $\alpha$}

In fact, $\alpha$ is closely related to the number of neighbours when we construct the graph. It would be easier to set the nearest neighbours number $k$ than tuning $\alpha$. Suppose we assign $k$ different neighbors to each data point, we can set $\gamma$ to be the average of $\{\alpha_i\}_{i=1}^N$. For any  $x_i$ , the Lagrangian function of problem (\ref{updates}) can be written as:
\begin{equation}
\mathcal{L}\left ( s_{i},\phi ,\varphi _{i} \right )=\frac{1}{2}\left \| s_{i}+\frac{1}{4\alpha _{i}\beta }d_{i} \right \|_{2}^{2}-\phi (s_{i}^{T}\mathbf{1}-1)-\varphi _{i}^{T}s_{i},
\end{equation}
where $\phi$, $\varphi_i\geq 0$ are Lagrangian multipliers. Applying KKT condition(Lemaréchal 2006), we get the optimal solution of $s_i$:
\begin{equation}
s_{ij}=(-\frac{d_{ij}}{4\alpha _{i}\beta }+\phi )_{+}.
\end{equation}
Considering the constraint $s_{i}^{T}\mathbf{1}=1$, we get
\begin{equation}\sum_{j=1}^{k}(-\frac{d_{ij}}{4\alpha _{i}\beta }+\phi )=1\\
\Rightarrow \phi =\frac{1}{k}+\frac{1}{4k\alpha _{i}\beta }\sum_{j=1}^{k}d_{ij}.
\label{aa}
\end{equation}
If the optimal $s_{i}$ has only $k$ nonzero elements, then $s_{i,k}> 0$ and $s_{i,k+1}= 0$. That is,
\begin{equation}\left\{\begin{matrix}
-\frac{d_{i,k}}{4\alpha_{i}\beta  }+\phi> 0, \\ 
-\frac{d_{i,k+1}}{4\alpha_{i}\beta  }+\phi\leq  0.
\end{matrix}\right.
\label{ab}
\end{equation}

Combining (\ref{aa}) and (\ref{ab}), we have the following inequality for $\alpha _{i}$:
\begin{equation}
\frac{1}{2\beta }(\frac{k}{2}d_{ik}-\frac{1}{2}\sum_{j=1}^{k}d_{ij})< \alpha _{i}\leq \frac{1}{2\beta }(\frac{k}{2}d_{i,k+1}-\frac{1}{2}\sum_{j=1}^{k}d_{ij}),
\end{equation}
where $d_{i1}, d_{i2}, \cdots, d_{iN}$ are sorted in ascending order.
We set $\alpha _{i}$ to its maximum and we set the overall $\alpha $ as the mean of $\alpha _{i}$. Then, the value of $\alpha $ becomes
\begin{equation}\alpha=\frac{1}{2N\beta }\sum_{i=1}^{N}(\frac{k}{2}d_{i,k+1}-\frac{1}{2}\sum_{j=1}^{k}d_{ij}).\end{equation}

With this equation, we can tune the value of $k$ instead of $\alpha $ since $k$ is an integer and has a small range, which is trivial to find an appropriate value.
\begin{table}[t]
\begin{center}
\caption{Statistics of datasets used in experiments} \label{tab:stat}
\renewcommand{\arraystretch}{1.2}
\renewcommand{\tabcolsep}{5mm}
\begin{tabular}{|c|c|c|c|}
  \hline
  Dataset & Dimension of Views & \# of Instances & \# of Class
  \\
  \hline
BBC & 4659/4633/4665/4684 & 145 & 2 \\
Sonar & 20/20/20  & 208 &2 \\
HW & 216/76/64/6/240/47 &2000 & 10\\
Reuters & 2000/2000/ 2000/ 2000/ 2000 & 1200 & 6\\
  \hline
\end{tabular}
\end{center}
\end{table}

\section{Experiments}

\subsection{Datasets}

We assess our approach on 4 publicly available datasets. We summarize the information of these datasets in Table \ref{tab:stat}.

\textbf{BBC} dataset is derived from the BBC sport website corresponding to sports news articles. Each document is split into four segments by separating the document into paragraphs. Each segment is at least 200 characters long and is logically associated with the original document from which it was obtained.\\

\textbf{Sonar} is a dataset with 208 observations on 60 variables. The features represent the energy within a particular frequency band, integrated over a certain period of time. There are two classes 0 if the object is a rock, and 1 if the object is a mine (metal cylinder). The 60 variables are divided into three parts, each representing a view of 20 variables.\\

\textbf{Handwritten numerals (HW)} dataset consists of 2000 data points evenly distributed in 0 to 9 digit classes with 200 data points in each class. We choose six types of features of all data points: profile correlations, Fourier coefficients of the character shapes, Karhunen-Love coefficients, morphological features, pixel averages in $2\times3$ windows, and Zernike moments.\\

\textbf{Reuters} is a textual dataset which is written in five different languages (English, French, German, Spanish, and Italian). All the documents are categorized into 6 classes. There are altogether five views and each of them represents documents written in one certain language. From each view, we randomly sample 1200 documents in a balanced manner, with each of the 6 classes having 200 documents.

\begin{table*}
	\begin{center}
	\renewcommand{\arraystretch}{1.2}
\renewcommand{\tabcolsep}{2mm}
		\caption{Semi-supervised classification accuracy (\%) on BBC} \label{tab:bbc}
		\begin{tabular}{|c|c|c|c|c|}
			\hline
			Dataset &\multicolumn{4}{c|}{BBC}
			\\
			\hline
			rate & 0.1 & 0.2 & 0.3 &0.5 \\
			\hline
			AMGL&49.50(2.68)&50.68(4.10)&49.22(3.97)&52.32(4.74)\\
			\hline
			MLAN&52.02(6.45)&55.73(3.43)&56.12(5.68)&57.40(5.28)\\
			\hline
			HFPF(1)&92.77(0.88)&92.24(0.97)&93.12(1.45)&92.64(1.92)\\
			\hline
			HFPF(2)&92.38(0.66)&92.76(1.10)&92.43(1.34)&92.15(2.71)\\
			\hline
			HFPF(3)&92.42(0.87)&92.72(1.49)&92.23(1.30)&93.47(2.78)\\
			\hline
		HFPF(4)&92.54(0.75)&92.67(1.17)&92.18(1.01)&92.00(1.85)\\
		\hline
		LMSSC &$\mathbf{92.86(1.22)} $ &$\mathbf{93.59(1.85)} $&$\mathbf{94.42(1.57)} $&$\mathbf{95.14(3.00)} $ \\
			\hline
		\end{tabular}
	\end{center}
\end{table*}

\subsection{Comparison Methods}

As a baseline, we apply the classic single-view semi-supervised classification method GFHF~\cite{zhu2003semi} to each view of the datasets. Moreover, we compare our method with two other popular multi-view semi-supervised classification methods: (a) Multi-view Learning with Adaptive Neighbours (MLAN)~\cite{nie2017multi}, (b) Auto-weighted Multiple Graph Learning (AMGL)~\cite{nie2016parameter}.
For each dataset, we randomly choose 10\%, 20\%, 30\%, and 50\% of points as labeled. This procedure is repeated 20 times. Finally, we report the average classification accuracy and deviation. For our proposed LMSSC\footnote{The code for our implementation is available: https://github.com/sckangz/LMVL}, all data points are assumed to have 15 nearest neighbours, i.e., we use $k=15$ to calculate $\alpha$.

\begin{table*}
	\begin{center}
	\renewcommand{\arraystretch}{1.2}
\renewcommand{\tabcolsep}{2mm}
		\caption{Semi-supervised classification accuracy (\%) on Handwritten} \label{tab:cap}
		\begin{tabular}{|c|c|c|c|c|}
			\hline
			Dataset &\multicolumn{4}{c|}{Handwritten}
			\\
			\hline
			rate & 0.1 & 0.2 & 0.3 &0.5 \\
			\hline
			AMGL&92.80(0.54)&93.85(0.58)&94.17(0.55)&94.89(0.62)\\
			\hline
			MLAN&$\mathbf{97.83(0.23)} $&$\mathbf{97.81(0.44)} $&$\mathbf{98.00(0.38)} $&98.10(0.41)\\
			\hline
			HFPF(1)&88.94(0.78)&91.28(0.77)&92.52(0.48)&93.12(0.44)\\
			\hline
			HFPF(2)&80.06(0.85)&81.90(0.99)&82.98(0.86)&83.39(1.05)\\
			\hline
			HFPF(3)&94.21(0.59)&95.79(0.38)&96.26(0.48)&96.59(0.47)\\
			\hline
			HFPF(4)&42.85(1.23)&44.88(0.90)&45.92(0.95)&47.62(1.30)\\
			\hline
			HFPF(5)&94.96(0.54)&96.60(0.44)&96.75(0.33)&97.25(0.44)\\
			\hline
			HFPF(6)&79.72(0.77)&82.03(0.62)&82.52(0.85)&82.93(0.88)\\
			\hline
			LMSSC&96.59(0.46)&97.10(0.38)&97.59(0.37) &$\mathbf{98.70(0.33)} $ \\
			\hline
		\end{tabular}
	\end{center}
\end{table*}

\subsection{Results}

All results are shown in Tables \ref{tab:bbc}-\ref{tab:sonar}. As expected, the accuracy of each method monotonously increases with the increase of label rate. We can see that the performance of HFPF heavily depends on the datasets and the specific view. On the other hand, our LMSSC consistently outputs high accuracy. This validates the advantage of multi-view learning. On BBC, Reuters, and Sonar datasets, our method outperforms AMGL and MLAN by a large margin. For HW dataset, our accuracy is comparable to MLAN method and outperforms others. Furthermore, we can see that AMGL and MLAN perform pretty well on Handwritten and Sonar, bad on BBC and Reuters. This instability is caused by the built graph which heavily depends on the quality of data. By contrast, LMSSC builds a graph on latent representation, which makes the graph robust to noise and outliers. 
\begin{table*}
	\begin{center}
		\caption{Semi-supervised classification accuracy (\%) on Reuters} \label{tab:reuters}
		\renewcommand{\arraystretch}{1.2}
\renewcommand{\tabcolsep}{2mm}
		\begin{tabular}{|c|c|c|c|c|}
			\hline
			Dataset &\multicolumn{4}{c|}{Reuters}
			\\
			\hline
			rate & 0.1 & 0.2 & 0.3 &0.5 \\
			\hline
			AMGL&35.44(4.95)&39.34(5.00)&42.70(3.48)&46.97(2.04)\\
			\hline
			MLAN&57.66(9.45)&63.83(5.83)&66.12(3.69)&70.66(1.77)\\
			\hline
			HFPF(1)&31.18(3.00)&35.22(3.76)&37.40(4.55)&41.13(4.79)\\
			\hline
			HFPF(2)&30.32(4.83)&33.50(3.35)&39.00(3.76)&42.29(1.92)\\
			\hline
			HFPF(3)&27.88(3.91)&35.76(2.86)&36.94(4.77)&39.70(3.93)\\
			\hline
			HFPF(4)&30.67(3.51)&36.03(2.74)&38.53(2.76)&42.18(2.29)\\
			\hline
			HFPF(5)&30.54(3.88)&35.38(1.94)&37.54(2.58)&41.08(1.96)\\
			\hline
			LMSSC &$\mathbf{60.66(2.52)} $&$\mathbf{68.08(2.74)} $&$\mathbf{76.48(1.97)} $& $\mathbf{87.28(1.38)} $ \\
			\hline
		\end{tabular}
	\end{center}
\end{table*}

\begin{table*}
\begin{center}
\caption{Semi-supervised classification accuracy (\%) on Sonar dataset} \label{tab:sonar}
\renewcommand{\arraystretch}{1.2}
\renewcommand{\tabcolsep}{2mm}
\begin{tabular}{|c|c|c|c|c|}
  \hline
  Dataset &\multicolumn{4}{c|}{Sonar}
  \\
  \hline
  rate & 0.1 & 0.2 & 0.3 &0.5 \\
\hline
AMGL&63.32(6.43)&71.85(3.44)&72.60(3.76)&74.95(5.97)\\
			\hline
			MLAN&61.78(4.93)&67.71(4.19)&69.11(3.80)&72.26(3.33)\\
			\hline
			HFPF(1)&66.50(4.35)&67.86(4.48)&69.17(3.60)&69.90(3.97)\\
			\hline
			HFPF(2)&55.51(4.00)&56.54(4.24)&57.97(4.19)&58.08(4.06)\\
			\hline
			HFPF(3)&57.03(4.68)&58.70(3.54)&59.72(3.69)&61.97(2.39)\\
			\hline
			LMSSC & $\mathbf{69.84(5.54)}$&$\mathbf{80.21(4.65)}$&$\mathbf{90.14(2.15)}$&$\mathbf{95.48(1.29)}$\\
  \hline
\end{tabular}
\end{center}
\end{table*}
\begin{figure*}[!htbp]
	\centering
	\subfigure[$r=5$]{
			\includegraphics[width=0.3\textwidth]{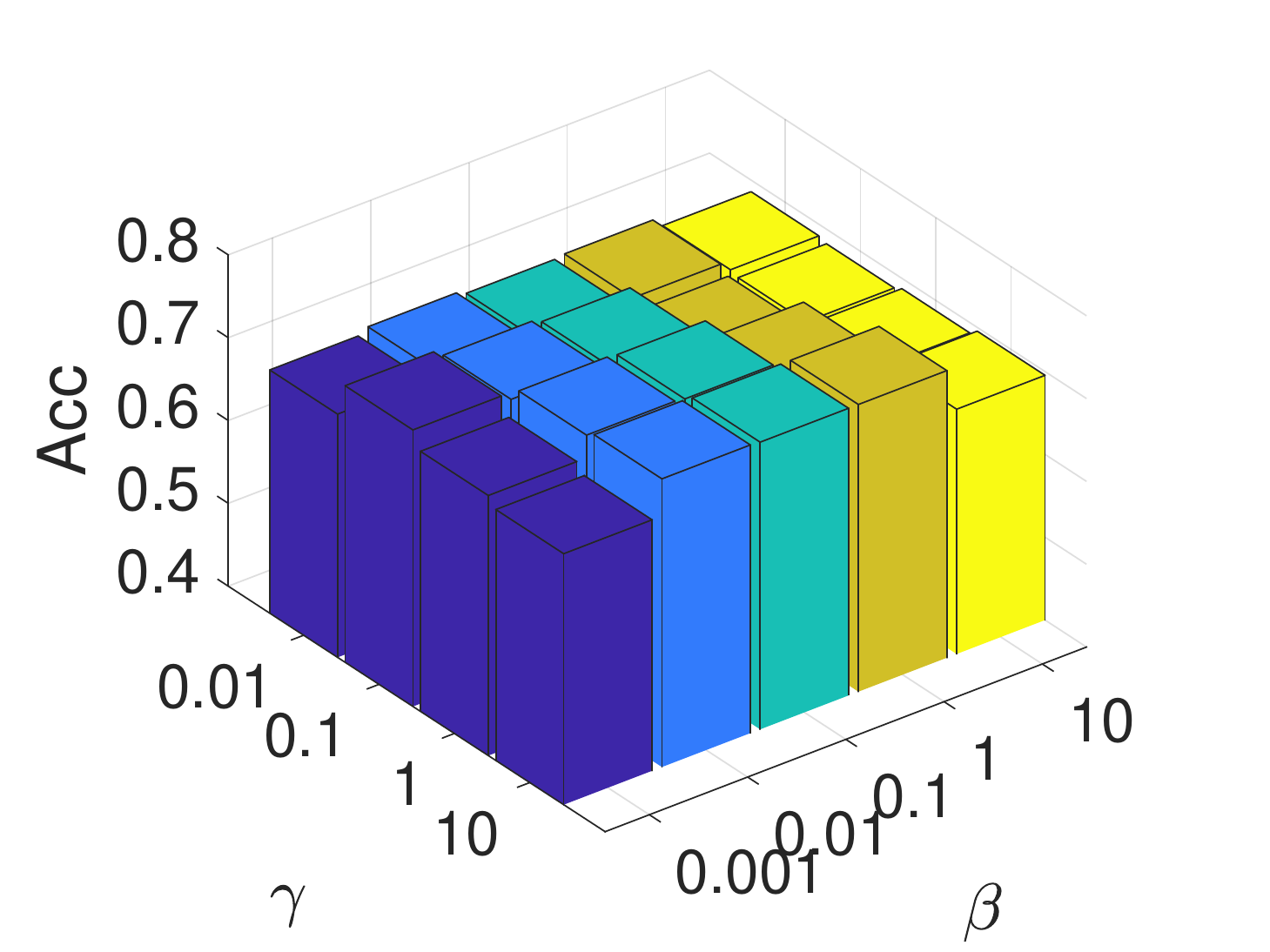}
	}
	\subfigure[$r=10$]{
			\includegraphics[width=0.3\textwidth]{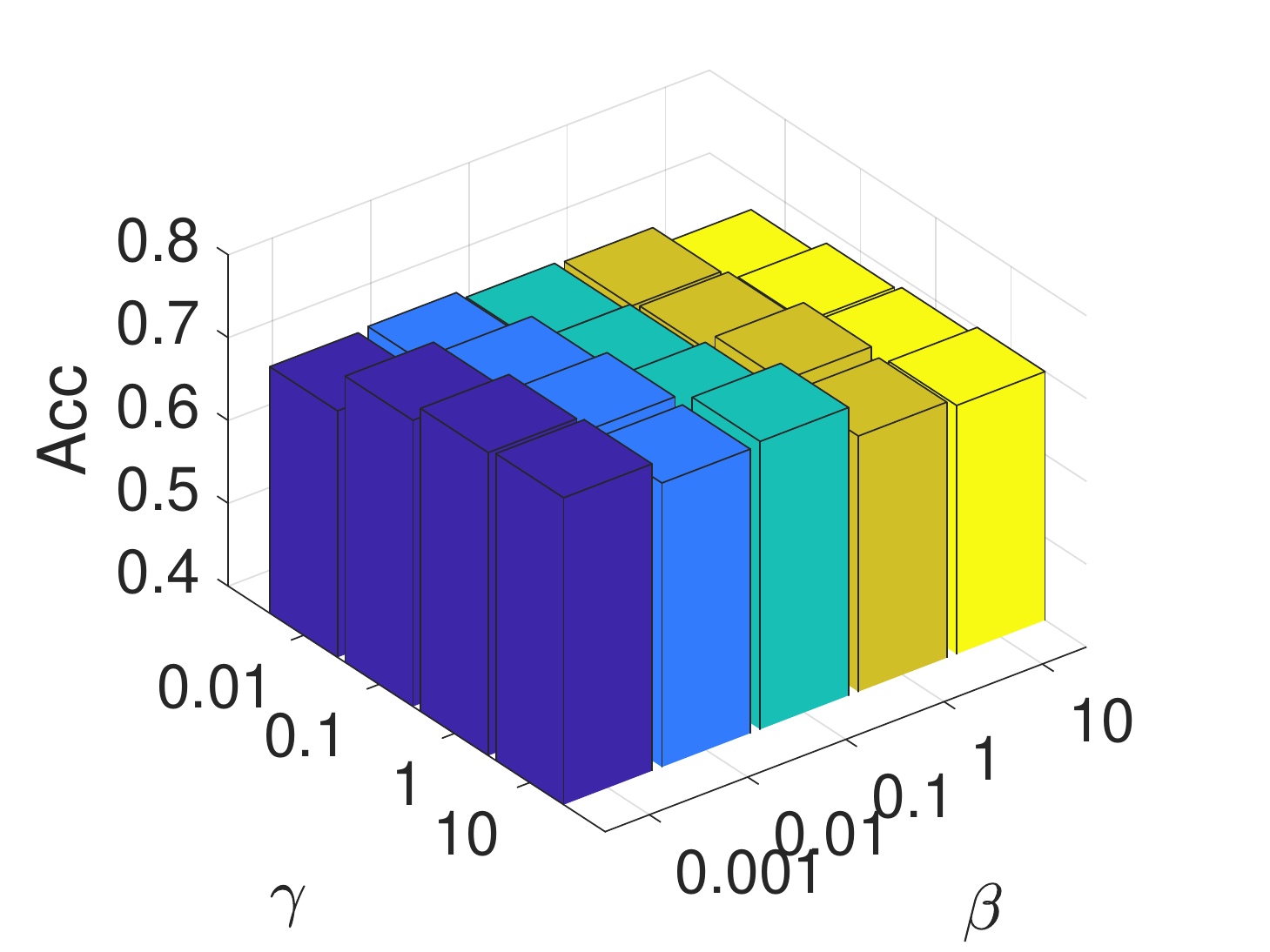}
	}%
	\subfigure[$r=15$]{
			\includegraphics[width=0.3\textwidth]{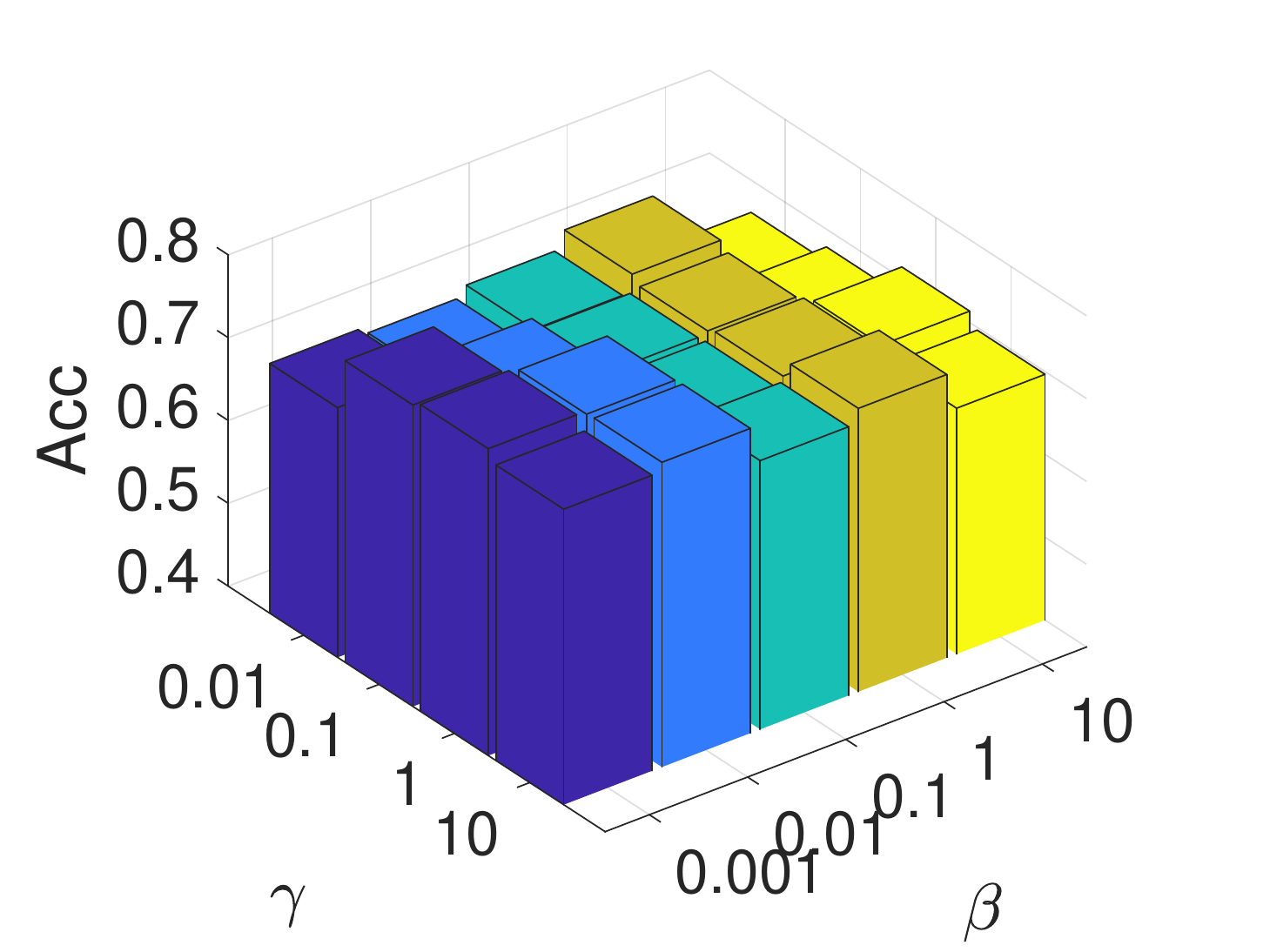}
	}%

	\caption{Parameter sensitivity for classification accuracy on Sonar dataset.}\label{sens}
\end{figure*}

\begin{figure*}[!htbp]
	\centering
	
		\includegraphics[width=0.8\textwidth]{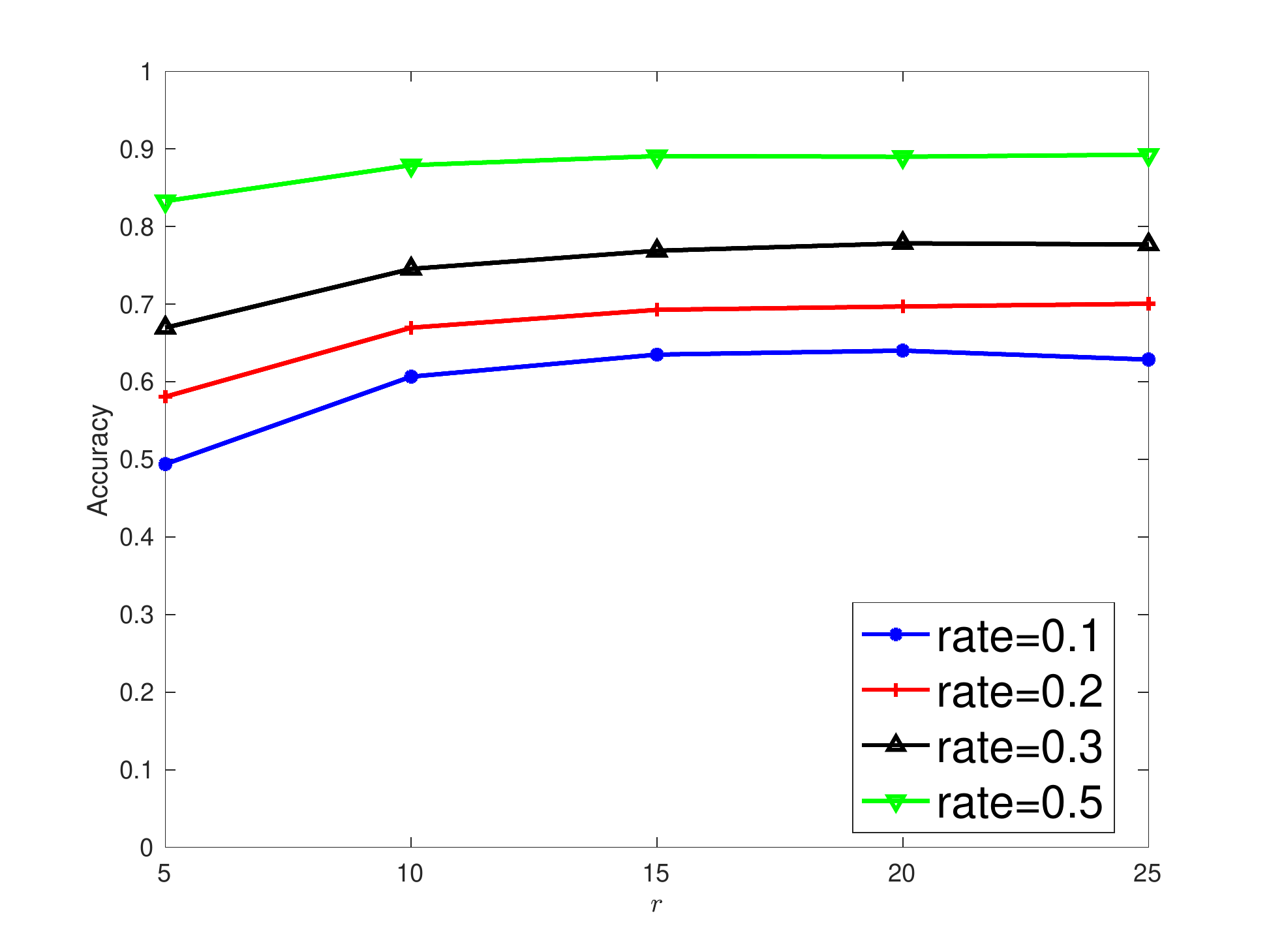}

	\caption{The influence of latent dimension $r$ on accuracy of Sonar dataset.  }
	\label{dim}
\end{figure*}
\subsection{Parameter Analysis}
In our model, there are altogether four parameters: $\alpha$, $\beta$, $\gamma$, and latent space dimension $r$. $\alpha$ is explicitly calculated. We use the Sonar dataset with 20\% label rate as an example to demonstrate the effects of $\beta$, $\gamma$, and $r$. As shown in Fig. \ref{sens}, accuracy changes slightly with a very wide range of parameters.

To explicitly demonstrate the influence of latent dimension, we vary the dimension $r$ from 5 to 25 with interval 5. For each $r$ value, we obtain its optimal accuracy. From Fig. \ref{dim}, we can see that after $r$ increases to 10, the performance becomes quite stable, which indicates that $H$ contains redundant information.
\section{Conclusion}
\label{sec:Con}

In this paper, we propose a novel multi-view semi-supervised classification method. It first learns a unique latent representation from original multi-view data. Based on this latent representation, a graph with structure guarantee is constructed using adaptive neighbours strategy. Eventually, latent representation learning, graph construction, and label prediction are seamlessly integrated together, which enjoys the benefits of joint optimization. Extensive experiments on real datasets verify the effectiveness of our proposed approach.

\acks{This paper was in part supported by Grants from the Natural
Science Foundation of China (Nos. 61806045 and
61572111), two Fundamental Research Fund for the Central
Universities of China (Nos. ZYGX2017KYQD177
and A03017023701012) and a 985 Project of UESTC (No.
A1098531023601041).}

\bibliography{ref}

%
%
%
%

\end{document}